\documentclass[runningheads]{llncs}
\usepackage{graphicx}
\begin{document}
\title{Clarifying the Half Full or Half Empty Question: Multimodal Container Classification\thanks{The authors gratefully acknowledge support from the
DFG (CML, MoReSpace, LeCAREbot), BMWK (SIDIMO, VERIKAS), and the European
Commission (TRAIL, TERAIS).}} 
\titlerunning{Multimodal Container Classification}
%
\author{Josua Spisak \and
Matthias Kerzel  \and
Stefan Wermter}
\authorrunning{J. Spisak et al.}
%
\institute{Knowledge Technology, Department of Informatics, University of Hamburg, Vogt-Koelln-Street 30, 22527 Hamburg, Germany
\url{http://www.knowledge-technology.info} \\
\email{\{josua.spisak\}@uni-hamburg.de}}
\maketitle              
\begin{abstract}
Multimodal integration is a key component of allowing robots to perceive the world. Multimodality comes with multiple challenges that have to be considered, such as how to integrate and fuse the data. In this paper, we compare different possibilities of fusing visual, tactile and proprioceptive data. The data is directly recorded on the NICOL robot in an experimental setup in which the robot has to classify containers and their content. Due to the different nature of the containers, the use of the modalities can wildly differ between the classes. We demonstrate the superiority of multimodal solutions in this use case and evaluate three fusion strategies that integrate the data at different time steps. We find that the accuracy of the best fusion strategy is 15\% higher than the best strategy using only one singular sense. 
\end{abstract}
%
%
%

%
%
%
%
%
%
\section{Introduction}
We constantly receive information and stimuli from all of our senses. Even in simple actions that we perform every day, such as drinking water, we will taste it, touch the bottle or glass holding the water, feel our muscles that help us lift it, and hear how we swallow it or how it moves around. We are processing all of these modalities at once, integrating their features \cite{zmigrod2013feature}. Robots, on the other hand, are often much more limited in their perception, often only singular senses are used for given tasks or a combination of just two or three senses. For these tasks, a limited diversity of senses is often enough, however, if we want robots to be able to freely interact with their environment, they need to be able to sense more of their environment. To facilitate this, we use three sensory modalities of our robot in this paper vision, tactile and proprioception and look at how to best integrate them. We do this on a task where perception with a single modality can be challenging. \\
\begin{figure}
    \centering
    \includegraphics[width=0.24\columnwidth]{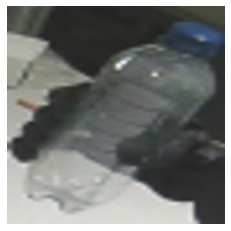}
    \includegraphics[width=0.24\columnwidth]{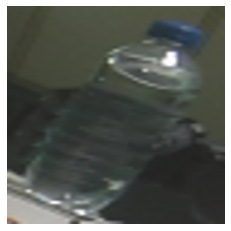}
    \includegraphics[width=0.24\columnwidth]{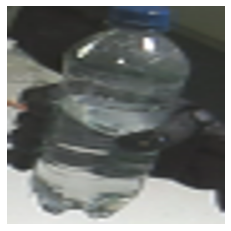}
    \includegraphics[width=0.24\columnwidth]{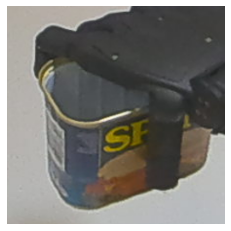}
    \includegraphics[width=0.24\columnwidth]{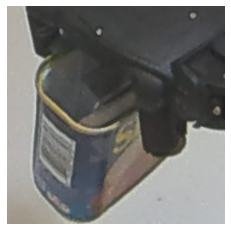}
    \includegraphics[width=0.24\columnwidth]{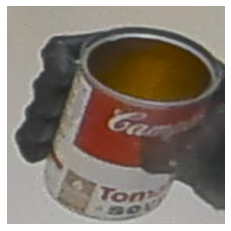}
    \includegraphics[width=0.24\columnwidth]{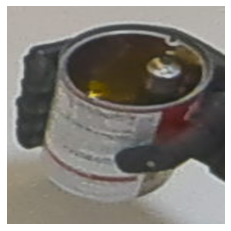}
    \caption{The seven object classes, from left to right, we have: the empty bottle, the filled bottle, the half filled bottle, the empty spam box, the filled spam box, the empty tomato can and the filled tomato can.}
    \label{fig:classes}
\end{figure}
In the theory of affordances \cite{gibson1977theory}, the perception of the world and specifically the perception of objects is discussed. The focus lies on perceiving what the world or objects can afford us. Depending on the observer, objects can have affordances, such as a ball affording to lift or to grasp if the size of the ball relative to the observer fits, if the ball is a bit bigger it might afford sitting on it or leaning against it and so on. The idea is that when we perceive anything we mainly perceive what it affords us, as that is what we need to perceive in order to interact with it. One group of objects that can be especially flexible when it comes to their affordances are containers. The content of containers can differ widely, changing what the container affords us. An empty container affords space to store things safely, while a container filled with water could afford drinking, washing or cooling. One of the challenges with creating robots that can learn from the world, imitate others, and use that learned knowledge to interact with the world, is that some objects are difficult to correctly perceive \cite{cui2022play}. This is a fundamental process on which further steps rely.\\
Robots tend to have multiple sensors and ways in which they can receive information about the world, so we need to look at how we can use that multimodal data and how to combine it. We want to explore which senses are useful and how to best integrate them to get a more complete concept of the world.
In this paper, we use a multimodal system to allow a robot to detect the contents of containers, thereby gaining a deeper understanding of the affordances of containers for the robot. This understanding allows the robot to interact with the world and learn from it, the robot can be helped by having a model of the world. Part of such a model should be the effects of actions. If a robot can understand the effect of actions, it can also predict what will happen after actions are performed. This can allow the robot to interact with others as it can understand the purpose or goal of actions. It can also allow it to learn from others \cite{montesano2008learning,lopes2007affordance}. As it can find actions that lead to the same effects as the actions it observes from others. The ability to detect the content of containers can also improve human-robot cooperation \cite{sciutti2018humanizing}.\\
Using multiple strategies for multimodal object recognition has been tried before \cite{castellini2011using,toprak2018evaluating}, however, to our knowledge it has never been used for the detection of content inside of containers, where the modalities encounter a higher number of challenges. The containers we use vary more than in other research studies. Where most studies only concern themselves with one type of container such as a glass and different contents \cite{guler2014s}, we experimented with three containers that differ in size, material and form.\\
The main contributions of this paper are summarised as follows:\\
\begin{itemize}
    \item By using a multimodal approach that utilises vision, touch and proprioception to detect the contents of containers, we improve the abilities of robots to interact with their environment and other actors.
    \item We use neural networks to classify objects perceived by our robot and compare three fusion strategies for multimodal integration on the same task.
    \item We collected multimodal data with a real-world humanoid robot of 3 objects each with different fill levels for a total of 7 classes. Examples of the seven classes can be seen in Figure \ref{fig:classes}.
\end{itemize}

\begin{figure}[t]
    \centering
    \includegraphics[width=1\textwidth]{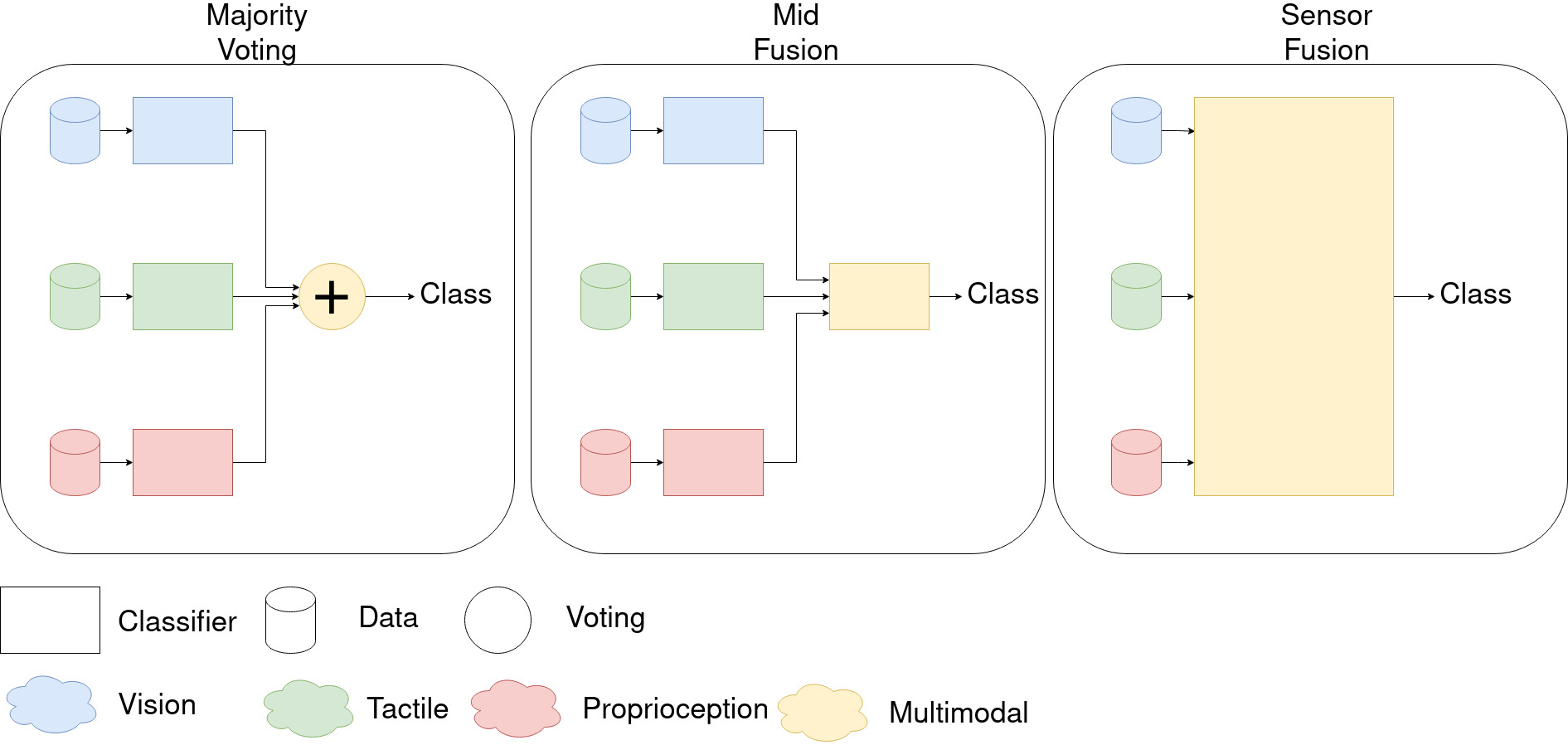}
    \caption{An overview of the three methods of multimodal fusion we use.}
    \label{fig:threeMethods}
\end{figure}
\section{Related work}
\subsection{Container content detection}
Detecting the content of containers can be done in many ways, often depending on the kind of container at hand. We can find some separation between them by looking at which modalities are used. Using vision to detect the content of containers is a common approach where it is possible to look into or through the container to see the content \cite{pau2020dataset,do2016probabilistic,pithadiya2011selecting}. The vision can consist of just RGB cameras or be improved further with specialised cameras such as depth cameras or CCD cameras. The perceived images can be processed with many kinds of mechanisms such as neural networks, edge detectors or probabilistic models. There are some containers where vision is not quite as useful to detect the content. This applies for instance to cans, where we cannot see anything inside them, or even fruits such as avocados whose colouring tells us little about their ripeness. A modality that can be used for these containers is the tactile sense \cite{chitta2010tactile}. Just like humans would touch an avocado to learn about its ripeness, machine learning mechanisms can use the data gathered by tactile sensors to classify what is inside of an object. Similarly, proprioception can be used to detect the weight of a container and extrapolate from that to information about the content \cite{piacenza2022pouring}.\\
To use the advantages of more than a single modality, multimodal approaches integrate senses to improve detection. While there is often an imbalance in the performance between senses, the combination of them does improve the overall results \cite{pieropan2014audio,jonetzko2020multimodal,hall1997introduction}.
\subsection{Multimodal integration} 
Of course, using multiple modalities means that we have to integrate them and bring them together. Fusion strategies are often separated into three groups. The three groups are known by multiple terms \cite{sanderson2004identity,ross2003information,lahat2015multimodal}, but tend to be quite similar. The first group of fusion strategies fuses the data gained from the sensors directly before any sort of mapping happens from the data to a desired result. This fusion can be done in many ways depending on the data and often challenges are encountered, such as different sizes of the data or other ways in which the data is hard to match between senses \cite{lahat2015multimodal}. Another group of fusion strategies only fuses the information at the very end of the process. Here, the senses are handled individually until they finish mapping from the information to the result, and the results from the senses are then fused to form a consensus. Some exemplary methods are majority voting, weighted majority voting, behaviour knowledge space, and Naive-Bayes classification \cite{mangai2010survey}. The last group is somewhat in between, here the data is first processed individually but fused before coming to clear results. The later we integrate the information, the easier this integration tends to be, with lesser training requirements or data requirements. The downside of later fusions is that it is easier to miss cross-modal interaction, as some information about how the modalities interact with each other can be lost \cite{turk2014multimodal}.

\section{Multimodal data set for container content classification} \label{section dataset}
We propose a multimodal data set including three modalities, vision, proprioception and tactile perception. We recorded this data set with the NICOL (Neuro-Inspired COLlaborator)\cite{kerzel2023nicol}. We started recording data from the cameras hosted in the head of NICOL, the effort of its joints and the data from the tactile sensors of the fingers. We put the container onto the table in front of NICOL, where the robot moves its arm to grasp the container. Then the robot lifts the container briefly, moving it around before putting it back in place. We filtered out the data where the robot was not holding the container for the tactile and proprioception data later on. The modalities are recorded at different frequencies depending on the robot's capabilities. We used three kinds of containers, a water bottle, and, from the YCB Object and Model set, the spam box as well as the tomato can. For the water bottle, we had three possible fill levels, empty, halfway filled and filled. The other objects could either be filled or empty. The bottle and the tomato can be grasped using a side grasp while we used a top grasp for the spam box. In total, we recorded around 77000 samples from the joints, 95000 samples from the tactile sensors and around 57000 images.\\

\begin{figure}[t]
\includegraphics[width=0.49\textwidth]{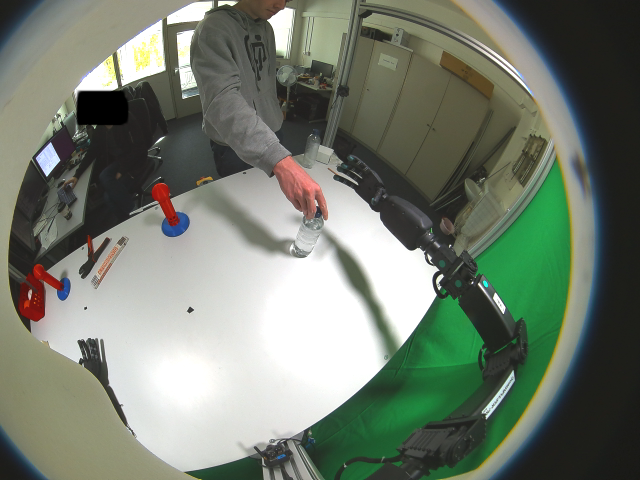}
\includegraphics[width=0.49\textwidth]{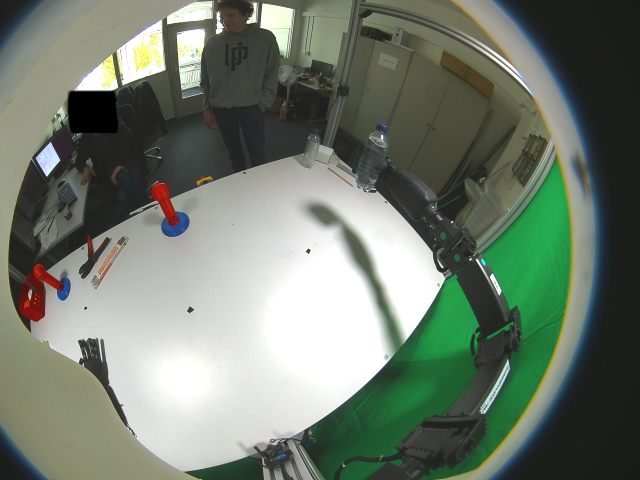}
\caption{Exemplary images recorded with the fish eye lens camera from the NICOL robot.} \label{fig1}
\end{figure}

For the vision, the robot recorded images at a resolution of 1920x1080 with a fish eye camera lens. We record 30 images per second. Which is quite high regarding the movement in the images. This means that the difference between two sequential images can be insignificant so we only used every tenth image. The containers only take up a small part of the image. Some exemplary images are shown in Figure \ref{fig1}. We annotated the data by hand, drawing bounding boxes around the containers and labelling them.\\
The tactile data was recorded from the tactile sensors that NICOL has in its five fingers. We used the values representing the directional forces relative to the fingertips, so the force that is measured along the x, the y and the z-axis. The tactile data is recorded at 50 samples per second. The tactile data does depend on the grip that the robot has on the container. Throughout the lifting attempts, the way each finger is positioned can change and impact the values we receive from the tactile sensors.\\
The proprioceptive data was gathered from the joints of NICOL. For each joint, we receive three values describing the position of the joint, the velocity of the joint and the effort of the joint. Similar to the tactile data, the proprioception data is impacted by the way in which the robot grasps the container. The proprioception data is recorded at a frequency of 40 samples per second. By proprioception data, we refer to the data we get from the joint motors of the robot, in our case, we have 23 joints.\\
Throughout the data collection, we used two ways of grasping the object, for the spam box we used a top grasp while we used a side grasp for the other objects.

\section{Approach}
We have used three ways of integrating our modalities, they are depicted in Figure \ref{fig:threeMethods}. The first strategy is majority voting, where we have individual classifiers for vision, proprioception and tactile data. We further have two voting mechanics to combine the output of these classifiers. The first one is hard voting, where we take the classification from each classifier as one value. So, this classification will either be filled, unfilled or half filled. Each classifier is counted as one voter. The option that was voted for by the majority of the classifiers is chosen. If there is no clear majority and there are multiple classifications with the highest number of votes, we declare the vote as undecided and count it as a false classification. Secondly, we have soft majority voting. Instead of taking a clear classification from each classifier, we use the whole output from each classifier. This means that each classifier provides us with an array that has seven values, one value for each possible classification. We add the arrays together, forming a new array also with three values. To form a final classification, we simply look at the highest of the three values and the class it correlates to.\\
Apart from the majority voting, we also used a neural network to fuse the modalities. We used the output from our classifiers as the input for another classifier, fusing the modalities and providing a classification. Finally, we fused the data at the beginning and created a NN that uses the fused data to directly classify the fill level. So, we have three strategies as to when the fusion happens in relation to the mapping from the data to the class.
\\

To allow any of this, we first have to synchronise the data. The modalities are recorded in different frequencies, and we want the samples to match for each of the modalities. To avoid having a sample multiple times, we took the vision data as our lead, as it had the lowest frequency. We then synchronised the data from all modalities so that the samples would match each other across the modalities. Like this, we compose a data set that has data from all modalities. This is the data set that we use in our experiments. This way, the individual classifiers are also looking at the same time frame.
\\
We preprocessed the visual data, inverted the colours, and cut down the image so that the container is a larger part of the image and in the centre of the image. We used the images with a resolution of 256x256 for our CNN, which worked as our visual classifier. The inversion of colour was done, so that parts of the images that are important to the detection of the content become easier to see \cite{pau2020dataset}. For example, the border where water meats air in the bottles. We cut down the image because the original images had a lot of background with only a small part being the container we wanted to focus on. This meant, that often the classifier had problems finding the container in the first place, not even getting to the task of identifying the contents.\\

In our mid fusion strategy, we create a classifier to fuse the output from the classifiers for each of the modalities. This classifier is a dense NN which consists of 4 layers with the activation function relu and an output layer with the sigmoid function and three neurons which provide three output values corresponding to the possible fill levels. We use Adam as our optimiser and the categorical cross-entropy loss function. We trained this network for 10 epochs with a batch size of 5. This strategy allows us to process more cross-modal information, as the classifier providing the final output has access to some information about each modality. It also means that it would still be possible to judge the quality of the individual modalities and gain information from them. This could be useful for evaluating the modalities or judging how trustworthy the results are.

\begin{figure}[t]
    \centering
    \includegraphics[width=1\textwidth]{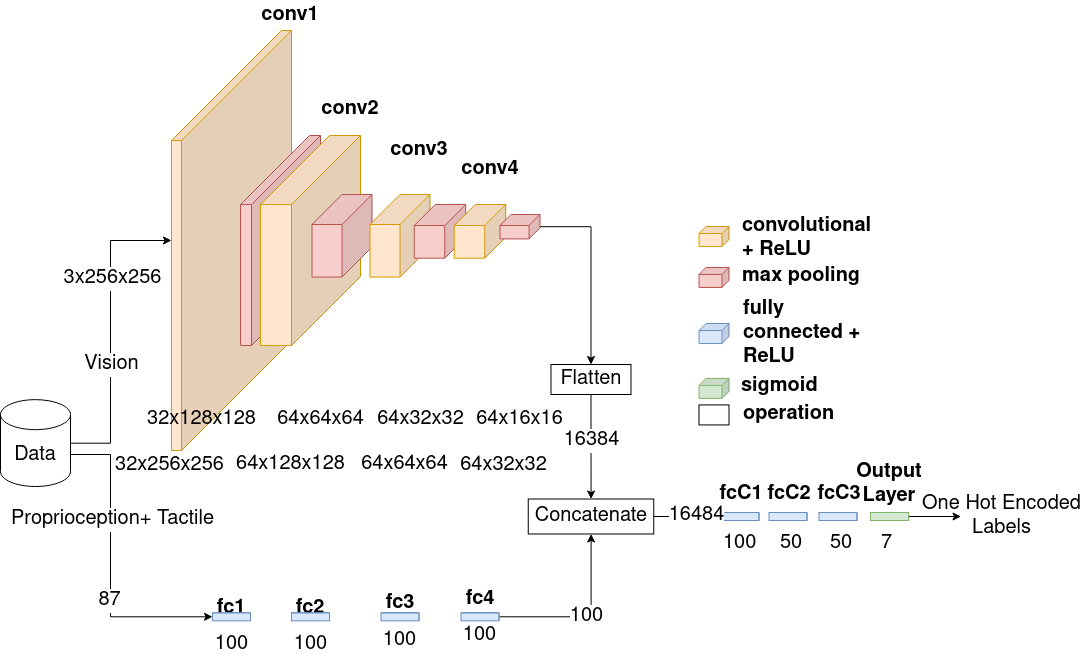}
    \caption{This figure displays the architecture of the neural network, that is utilised in our sensor fusion strategy. The sensory data, gathered from the robot's vision, joints and tactile sensors, is used as the input. The outputs are one-hot encoded labels corresponding to the seven possible classes we have. The number below the layers denotes their output shape.}
    \label{fig:SensorFusionArchitecture}
\end{figure}
In the sensor fusion strategy, the classifier we create needs to be able to process both visual data as well as data from the tactile sensors or the joints. To facilitate this, there are two input arms, one consisting of convolutional layers while the other one uses dense layers. The convolutional arm has 4 convolutional blocks of a convolutional layer and a max pooling layer, each before a dense layer. The dense arm has four dense layers with 100 neurons each. Both arms then lead into a dense layer with 100 neurons, followed by three more dense layers before the output layer with three neurons and the sigmoid activation function which produced our output. The model can be seen in Figure \ref{fig:SensorFusionArchitecture}.
The output is the same as with the other models and has one value correlating to each possible fill level. 
This strategy allows us to use a lot of cross-modal information. The model has direct sensory data from each of the modalities and can gain a deeper understanding of how they relate to each other and the desired output. However, the model also has to deal with more complex data, so its task is more difficult. We also lose some understanding about the model itself, as it would be a lot harder to tell afterwards how important which modality was to the result and whether or not we need all modalities or how to improve the data collection. \\

To gain a deeper understanding of each of our modalities, we further researched how classifiers, that only depend on one of the modalities, perform. Starting with the vision, we used the classifier that is also used for the majority voting. We have a CNN with eight convolutional blocks, each consisting of one convolutional layer with a relu activation function and max pooling layer. After the convolutional blocks, we have one dense layer with 64 neurons which also uses the relu activation function. Finally, we have our output layer which has three neurons and uses the sigmoid activation function to produce our output which consists of three values, each corresponding to one possible fill level of the container. We used the Adam optimiser and the categorical cross-entropy loss function. We trained the network for 10 epochs and used an 80:20 split for the training and validation set. The training data is a subset of the data set introduced in section \ref{section dataset}.\\
For the other modalities, we used a dense neural network consisting of four dense layers with the relu activation function and one output layer with three neurons which has the sigmoid function as its activation function. The only difference between the tactile classifier and the proprioception classifier is, that the tactile classifier has 15 input values which are the force directions along the x, y and z axis for each of the five fingers while the proprioception model has 69 input values which are the velocity, position and effort of each of the 23 joints. Both of these models use the Adam optimiser and the categorical cross-entropy loss function. We did, however, find that the proprioceptive model needs significantly more epochs to converge than the tactile model; therefore, the tactile model is trained for 10 epochs while the proprioception model is trained for 50 epochs.\\



\section{Results and Discussion}
\begin{table}[]
\caption{Comparison of the classifiers on the synchronised data from ten training runs.}

    \centering
    \begin{tabular}{|c|c|c|}
    \hline
         Classification Method & Validation Accuracy&  Validation Standard Deviation  \\
         \hline
         Tactile               & 74.9\% & 2.2\% \\
         Proprioception        & 43.3\% & 1.4\% \\
         Vision                & 60.9\% & 8.2\% \\
         Hard Majority Voting  & 65.4\% & 5.8\% \\
         Soft Majority Voting  & 83.5\% & 4.9\% \\
         Mid Fusion         & \textbf{90.6\%} & 3.7\% \\
         Sensor Fusion          & 82.8\% & 6.8\% \\
         \hline
    \end{tabular}
    
    \label{tab:ClassifierResultsComparison}
\end{table}

Table \ref{tab:ClassifierResultsComparison} shows the results of our fusion strategies, as well as the results of using only singular modalities. The results are averaged from ten training runs. The data used changed in between runs but stayed the same in one run for each classification method. The only way of fusing our data that performs worse than any of the individual modalities is the hard majority fusion. Here the negative impact of the more inaccurate modalities can be seen. As the other fusion strategies use more sophisticated approaches to fuse the data, they can find more nuances and be less impacted by the less accurate modalities. With most fusion strategies performing better than even the best modality, we show the benefits of multimodality. The best-performing fusion strategy is the Mid Fusion, in which we created a NN which takes the output of the individual classifiers as its input and produces seven values as its output which correlate to the seven possible classes. While this method keeps less cross-modal information than the sensor fusion strategy which directly takes the data from all sensors as its input, the data that needs to be processed is also less complex. This advantage turns out to be more important than the additional information. The soft majority voting and the sensor fusion strategy perform similarly, and both still outperform the best modality by more than 0.07 accuracy. \\

Out of the individual modalities, the classifier using the tactile data performs the best, followed by the classifier using the visual data and finally the classifier using the proprioception data. The tactile sensors can tell us a lot about the weight of the object, and the different forms of the object also make it so that they are grasped differently. Combining this information with the tactile data allows the classifier to be able to accurately differentiate between the classes. For the visual data, there are more obstacles to overcome, for some of the recorded samples it is almost impossible to tell whether a given container is filled or not with the hand of the robot occluding the content of the container. Another hindrance is the transparent nature of the bottles and the water they are filled with which can also make it hard to detect the fill level. These factors lead to the visual data, leading to a less accurate classifier. The classifier using the proprioception data had the lowest accuracy. While it was able to have a far better idea about the container than could be gained from simply guessing, the data from the motors is quite noisy, especially with the grasping and lifting process differing with each attempt.\\

    
\begin{figure}[t]
\centering
    \includegraphics[width=0.8\columnwidth]{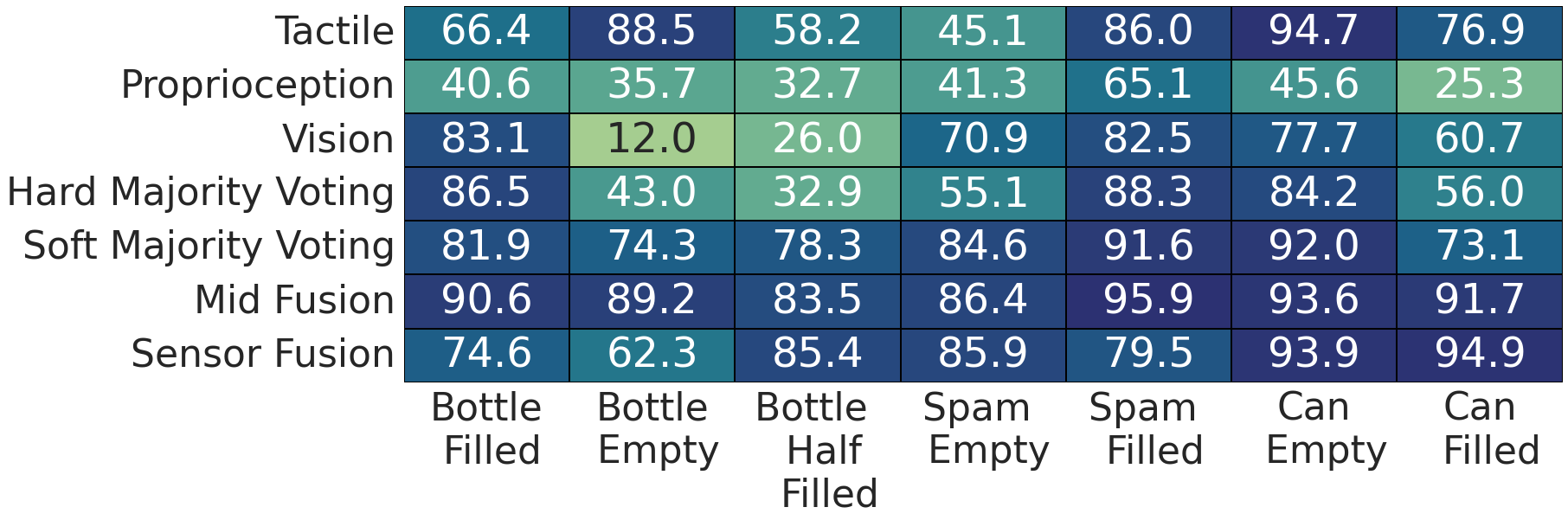}
\caption{Comparison of the average results of the Classifiers per class over ten runs in percent.}
\label{fig:ClassifierResultsComparisonByClass}
\end{figure}

\begin{figure}[t]
\centering
    \includegraphics[width=0.8\columnwidth]{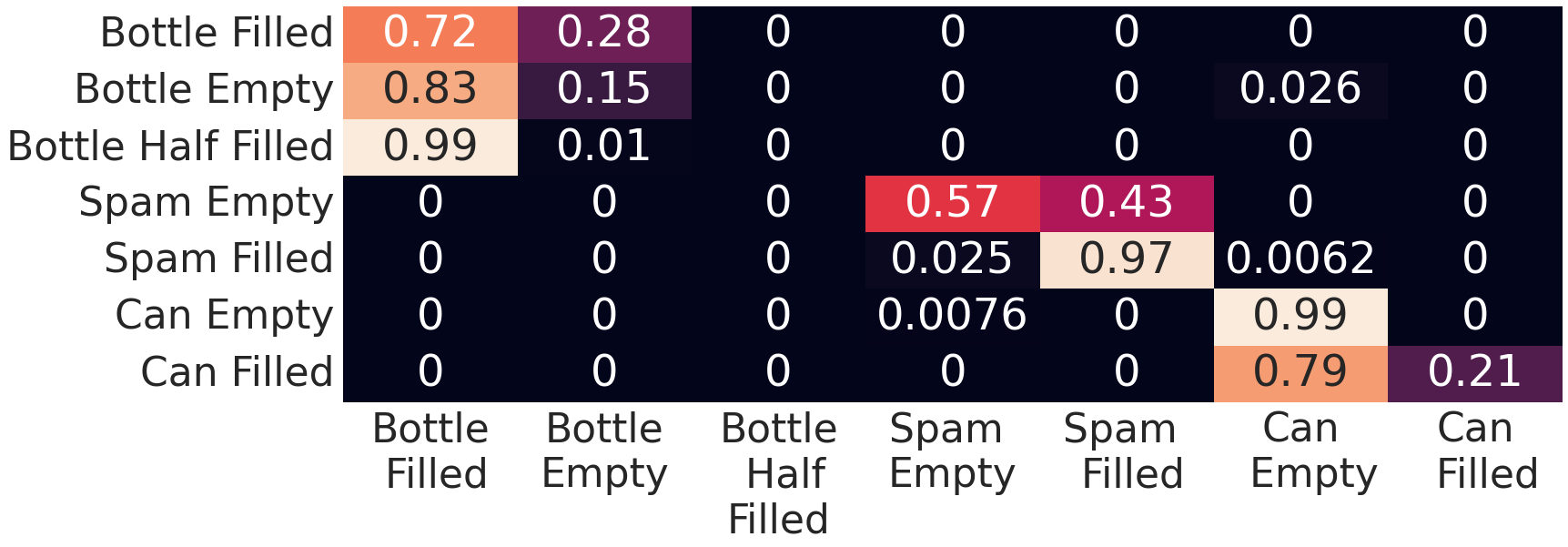}
\caption{Confusion Matrix for the vision classifier from a single run.}
\label{fig:confMats}
\end{figure}
While each of the modalities has its challenges, we have already seen that the combination of them improves the results, to gain a deeper understanding of the benefit that the combination of modalities provides, we compare how each classification method performs for each class in Table \ref{fig:ClassifierResultsComparisonByClass}. The results in the table are averaged over ten runs. The first thing that comes to note is that all of the best results for the classes come either from the mid fusion strategy, the sensor fusion strategy or, in one case, the Classifier using only tactile data. While the mid fusion strategy is not surprising here as we had already seen in Table \ref{tab:ClassifierResultsComparison} that it produced the best results overall, it is unexpected that for two of the seven classes, the sensor fusion strategy is the most accurate. The soft majority voting had a better overall accuracy, but also ends up being less volatile, having higher lows and lower highs than the sensor fusion strategy. With the main difference between the fusion strategies being the amount of cross-modal information available, it appears that for the classes ``Bottle Half Filled" and ``Can Full", this information is more important than for other classes. The other possible explanation is that the sensor fusion strategy does not make use of the classifiers for the individual modalities. So, it could be possible that the mid fusion and the soft majority voting perform worse for these three classes than the sensor fusion because the individual classifiers are less sure about these classes, so they could have high values for the class as well as a secondary class. With the class ``Bottle Half Filled", we can easily imagine that the classifiers see a larger similarity to the classes ``Bottle Filled" and ``Bottle Empty" than they see between these two classes directly. Similarly, the classes ``Can Empty" and ``Can Full" could be quite close to each other as they do look quite similar, and the container is less prone to be deformed during the grasp regardless of its fill level.\\

We can also see some peculiarities with individual modalities, The visual classifier seems to be particularly challenged by the classes ``Bottle Empty" and ``Bottle Half Filled". If we look further at the individual runs, we can see that there is always one bottle label that performs very well, most of the time that is the ``Bottle Filled" but on some runs, it is one of the others, the problem seems to be to differentiate between the fill levels of bottles and not in detecting the bottle. We can also see this in the confusion matrix shown in Figure \ref{fig:confMats}, where the classifier has a strong bias towards the ``Bottle Filled" class whenever any of the bottle classes should be predicted. The tactile data finds the class ``Spam Empty" to be the hardest to detect. Of course, the tactile data also provides the best result for the class ``Can Empty". That the visual classifier has such difficulties with two of the bottle classes could be explained by the challenge of transparency, why the tactile sensors performed worse on the ``Spam Empty" class would need to be researched further.

\section{Conclusion}
In this paper, we present multiple ways of classifying containers and their content, integrating up to three modalities. Our experiments compare different fusion strategies and showcase their strengths and weaknesses on data collected by a new robot. We evaluate the results and find that the best-performing fusion strategy utilises a NN to combine the results of individual classifiers for each of the modalities. We find a NN that can accurately classify multiple containers and their content which improves the ability of the robot to perceive the world and learn about the objects in said world, which is necessary to discern self from others. The large variance in shape appearance and material of the containers in our data sets lets us find strengths and weaknesses of the sensory modalities as well as how we can overcome them by fusing the modalities. Future work can include expanding the data set with more containers.
\bibliographystyle{splncs04}
\bibliography{mybib}

\begin{thebibliography}{10}
\providecommand{\url}[1]{\texttt{#1}}
\providecommand{\urlprefix}{URL }
\providecommand{\doi}[1]{https://doi.org/#1}

\bibitem{castellini2011using}
Castellini, C., Tommasi, T., Noceti, N., Odone, F., Caputo, B.: Using object
  affordances to improve object recognition. IEEE transactions on autonomous
  mental development  \textbf{3}(3),  207--215 (2011)

\bibitem{chitta2010tactile}
Chitta, S., Piccoli, M., Sturm, J.: Tactile object class and internal state
  recognition for mobile manipulation. In: 2010 IEEE International Conference
  on Robotics and Automation. pp. 2342--2348. IEEE (2010)

\bibitem{cui2022play}
Cui, Z.J., Wang, Y., Shafiullah, N.M.M., Pinto, L.: From play to policy:
  Conditional behavior generation from uncurated robot data. arXiv preprint
  arXiv:2210.10047  (2022)

\bibitem{do2016probabilistic}
Do, C., Schubert, T., Burgard, W.: A probabilistic approach to liquid level
  detection in cups using an rgb-d camera. In: 2016 IEEE/RSJ International
  Conference on Intelligent Robots and Systems (IROS). pp. 2075--2080. IEEE
  (2016)

\bibitem{gibson1977theory}
Gibson, J.J.: The theory of affordances. Hilldale, USA  \textbf{1}(2),  67--82
  (1977)

\bibitem{guler2014s}
G{\"u}ler, P., Bekiroglu, Y., Gratal, X., Pauwels, K., Kragic, D.: What's in
  the container? classifying object contents from vision and touch. In: 2014
  IEEE/RSJ International Conference on Intelligent Robots and Systems. pp.
  3961--3968. IEEE (2014)

\bibitem{hall1997introduction}
Hall, D.L., Llinas, J.: An introduction to multisensor data fusion. Proceedings
  of the IEEE  \textbf{85}(1),  6--23 (1997)

\bibitem{jonetzko2020multimodal}
Jonetzko, Y., Fiedler, N., Eppe, M., Zhang, J.: Multimodal object analysis with
  auditory and tactile sensing using recurrent neural networks. In:
  International Conference on Cognitive Systems and Signal Processing. pp.
  253--265. Springer (2020)

\bibitem{kerzel2023nicol}
Kerzel, M., Allgeuer, P., Strahl, E., Frick, N., Habekost, J.G., Eppe, M.,
  Wermter, S.: Nicol: A neuro-inspired collaborative semi-humanoid robot that
  bridges social interaction and reliable manipulation. arXiv preprint
  arXiv:2305.08528  (2023)

\bibitem{lahat2015multimodal}
Lahat, D., Adali, T., Jutten, C.: Multimodal data fusion: an overview of
  methods, challenges, and prospects. Proceedings of the IEEE  \textbf{103}(9),
   1449--1477 (2015)

\bibitem{lopes2007affordance}
Lopes, M., Melo, F.S., Montesano, L.: Affordance-based imitation learning in
  robots. In: 2007 IEEE/RSJ international conference on intelligent robots and
  systems. pp. 1015--1021. IEEE (2007)

\bibitem{mangai2010survey}
Mangai, U.G., Samanta, S., Das, S., Chowdhury, P.R.: A survey of decision
  fusion and feature fusion strategies for pattern classification. IETE
  Technical review  \textbf{27}(4),  293--307 (2010)

\bibitem{montesano2008learning}
Montesano, L., Lopes, M., Bernardino, A., Santos-Victor, J.: Learning object
  affordances: from sensory--motor coordination to imitation. IEEE Transactions
  on Robotics  \textbf{24}(1),  15--26 (2008)

\bibitem{pau2020dataset}
Pau, D., Kumar, B.P., Namekar, P., Dhande, G., Simonetta, L.: Dataset of sodium
  chloride sterile liquid in bottles for intravenous administration and fill
  level monitoring. Data in Brief  \textbf{33},  106472 (2020)

\bibitem{piacenza2022pouring}
Piacenza, P., Lee, D., Isler, V.: Pouring by feel: An analysis of tactile and
  proprioceptive sensing for accurate pouring. In: 2022 International
  Conference on Robotics and Automation (ICRA). pp. 10248--10254. IEEE (2022)

\bibitem{pieropan2014audio}
Pieropan, A., Salvi, G., Pauwels, K., Kjellstr{\"o}m, H.: Audio-visual
  classification and detection of human manipulation actions. In: 2014 IEEE/RSJ
  International Conference on Intelligent Robots and Systems. pp. 3045--3052.
  IEEE (2014)

\bibitem{pithadiya2011selecting}
Pithadiya, K.J., Modi, C.K., Chauhan, J.D.: Selecting the most favourable edge
  detection technique for liquid level inspection in bottles. International
  Journal of Computer Information Systems and Industrial Management
  Applications (IJCISIM) ISSN pp. 2150--7988 (2011)

\bibitem{ross2003information}
Ross, A., Jain, A.: Information fusion in biometrics. Pattern recognition
  letters  \textbf{24}(13),  2115--2125 (2003)

\bibitem{sanderson2004identity}
Sanderson, C., Paliwal, K.K.: Identity verification using speech and face
  information. Digital Signal Processing  \textbf{14}(5),  449--480 (2004)

\bibitem{sciutti2018humanizing}
Sciutti, A., Mara, M., Tagliasco, V., Sandini, G.: Humanizing human-robot
  interaction: On the importance of mutual understanding. IEEE Technology and
  Society Magazine  \textbf{37}(1),  22--29 (2018)

\bibitem{toprak2018evaluating}
Toprak, S., Navarro-Guerrero, N., Wermter, S.: Evaluating integration
  strategies for visuo-haptic object recognition. Cognitive computation
  \textbf{10},  408--425 (2018)

\bibitem{turk2014multimodal}
Turk, M.: Multimodal interaction: A review. Pattern recognition letters
  \textbf{36},  189--195 (2014)

\bibitem{zmigrod2013feature}
Zmigrod, S., Hommel, B.: Feature integration across multimodal perception and
  action: a review. Multisensory research  \textbf{26}(1-2),  143--157 (2013)

\end{thebibliography}




\end{document}